# WaterCopilot: An AI-Driven Virtual Assistant for Water Management

Keerththanan Vickneswaran
Water Futures Data & Analytics
International Water Management
Institute
Sri Lanka
K.Vickneswaran@cgiar.org

Mariangel Garcia Andarcia
Water Futures Data & Analytics
International Water Management
Institute
Sri Lanka
M.GarciaAndarcia@cgiar.org

Hugo Retief
Water Futures Data & Analytics
Association for Water and Rural
Development
South Africa
H.Retief@cgiar.org

Chris Dickens
Water Futures Data & Analytics
International Water Management
Institute
Sri Lanka
C.Dickens@cgiar.org

Paulo Silva
Water Futures Data & Analytics
International Water Management
Institute
Sri Lanka
P.Silva@cgiar.org

*Sustainable water resource management in transboundary river basins is challenged by fragmented data, limited real-time access, and the complexity of integrating diverse information sources. This paper presents WaterCopilot—an AI-driven virtual assistant developed through collaboration between the International Water Management Institute (IWMI) and Microsoft Research for the Limpopo River Basin (LRB) to bridge these gaps through a unified, interactive platform. Built on Retrieval-Augmented Generation (RAG) and tool-calling architectures, WaterCopilot integrates static policy documents and real-time hydrological data via two custom plugins: the iwmi-doc-plugin, which enables semantic search over indexed documents using Azure AI Search, and the iwmi-api-plugin, which queries live databases to deliver dynamic insights such as environmental-flow alerts, rainfall trends, reservoir levels, water accounting, and irrigation data. The system features guided multilingual interactions (English, Portuguese, French), transparent source referencing, automated calculations, and visualization capabilities. Evaluated using the RAGAS framework, WaterCopilot achieves an overall score of 0.8043, with high answer relevancy (0.8571) and context precision (0.8009). Key innovations include automated threshold-based alerts, integration with the LRB Digital Twin, and a scalable deployment pipeline hosted on AWS. While limitations in processing non-English technical documents and API latency remain, WaterCopilot establishes a replicable AI-augmented framework for enhancing water governance in data-scarce, transboundary contexts. The study demonstrates the potential of this AI assistant to support informed, timely decision-making and strengthen water security in complex river basins.*

*Keywords: Water resource management, Retrieval-Augmented Generation (RAG), Limpopo River Basin, Azure AI, Real-time APIs, Multilingual chatbots, Digital Twin, AWS deployment, RAGAS evaluation*

I. INTRODUCTION

The rapid evolution of Artificial Intelligence (AI) and Natural Language Processing (NLP) has greatly enhanced the capabilities of chatbots, enabling them to address complex, domain-specific challenges effectively. Large Language Models (LLMs), such as Generative Pre-trained Transformer (GPT)-4, have demonstrated exceptional abilities in interpreting user queries, retrieving relevant information, and generating contextually accurate responses. However, LLMs are inherently constrained by their reliance on static training data, limiting their capacity to deliver real-time, dynamic insights in specialized fields [1]. To overcome this limitation, Retrieval-Augmented Generation (RAG) techniques and tool-calling mechanisms have emerged as transformative approaches [2]. By enabling LLMs to access external, real-time, and static data sources through dynamic tool-based interactions, these strategies bridge the gap between general-purpose knowledge and domain-specific expertise.

Despite the significant potential of these technologies, their application in the field of Water Resource Management (WRM), particularly for complex, transboundary river systems like the LRB, remains underexplored. The LRB is a vital transboundary watershed spanning Botswana, Mozambique, South Africa, and Zimbabwe, supporting regional agriculture, biodiversity, and water security [3]. The basin faces significant challenges in managing its water resources due to fragmented data sources, outdated manual processing workflows, and limited real-time monitoring capabilities. Accessing accurate, timely, and comprehensive environmental information, such as rainfall patterns, river flows, water availability, and environmental flow thresholds, is critical for policymakers, researchers, and water managers, yet remains difficult with traditional systems [4].

To address these gaps, this research proposes the development of WaterCopilot, an AI-driven virtual assistant designed to provide an integrated, user-friendly, and real-time platform for water information retrieval and decision support across the LRB. The system builds upon the Digital Twin concept for the LRB, which integrates real-time data, simulation models, and reasoning tools to create a virtual representation of the basin [5]. WaterCopilot transforms traditional water management workflows by offering a conversational interface that facilitates seamless access to updated environmental data, static reports, and predictive insights. By leveraging Generative AI technologies, including Retrieval-Augmented Generation (RAG), Azure OpenAI Services, and real-time API integrations---including data sourced from operational hydrological models like SWAT+ [6], [7]---WaterCopilot aims to empower users with an efficient and personalized tool that not only simplifies information retrieval but also promotes informed and sustainable water management decisions across a highly diverse and critically important river basin.

*A. Research Objectives*

This research aims to achieve several key objectives, with a primary focus on improving the accessibility, reliability, and contextual relevance of water resource management systems in the Limpopo River Basin through innovative applications of LLMs, RAG, function calling and AI-driven virtual assistants:

- Design an AI-driven platform capable of providing comprehensive, real-time, and historical water resource information through WaterCopilot, which centralizes access to critical environmental data including rainfall patterns, river flows, water availability, environmental flow thresholds, and policy documents

- Evaluate the reliability and trustworthiness of LLMs integrated with dynamic retrieval mechanisms for domain-specific applications in water management

- Assess WaterCopilot's ability to deliver accurate, context-aware, and source-referenced responses to user queries in the transboundary Limpopo River Basin context

## II. RELATED WORK

LLMs represent a pivotal advancement in NLP, enabling high accuracy across a wide range of linguistic tasks. These models are primarily based on the Transformer architecture introduced by Vaswani et al. [8], which utilizes a multi-head self-attention mechanism to effectively capture long-range contextual dependencies within text. Modern LLMs are typically pre-trained on large-scale unlabeled corpora using self-supervised learning objectives, allowing them to learn rich syntactic and semantic representations of language. Based on their architectural design and training objectives, LLMs can be broadly categorized into three types: autoregressive models (e.g., GPT), which predict the next token in a sequence; masked language models (e.g., BERT), which infer masked tokens using bidirectional context; and encoder–decoder models (e.g., T5), which are optimized for sequence-to-sequence tasks such as translation and summarization. These pre-trained representations can subsequently be fine-tuned or utilized via in-context learning for specific downstream applications, leading to significant performance improvements across diverse NLP tasks [9]

*A. Applications of Pre-trained LLMs*

Pre-trained LLMs have been successfully applied to a wide range of NLP tasks. GPT models excel in text generation, enabling coherent and context-aware conversational AI systems [10]. BERT has been used extensively in tasks such as question answering (e.g., SQuAD datasets) and sentiment classification, leveraging its ability to generate bidirectional embeddings to understand nuanced semantic relationships [11]. Similarly, encoder-decoder models like T5 have been employed for text summarization, machine translation, and more complex multi-task scenarios, showcasing their versatility in "text-in, text-out" transformations [12].

*B. Challenges in Performance of LLMs*

Despite their capabilities, LLMs face critical challenges that limit their robustness and reliability. One prominent issue is shortcut learning, where models rely on superficial correlations or biases in the training data rather than developing genuine semantic understanding. For instance, in reading comprehension tasks, LLMs have been observed to predict answers based on lexical overlap between the question and context, ignoring the deeper meaning of the input [13]. Similarly, positional and stylistic biases often lead to incorrect predictions in question-answering and entailment tasks. These shortcuts undermine generalization to out-of-distribution data and make models vulnerable to adversarial inputs. Moreover, the current purely data-driven paradigm of training LLMs, while effective for large-scale corpora, falls short for tasks requiring contextual reasoning or domain-specific knowledge. Addressing these issues often necessitates extensive fine-tuning, which is computationally expensive and may still fail to eliminate inherent biases in the data [13].

The phenomenon of hallucination—where LLMs generate plausible yet factually incorrect outputs—is fundamentally rooted in the computational constraints of learning theory. Formal proofs indicate that hallucinations are an unavoidable consequence of model architecture, defined specifically as inconsistencies between generated outputs and a computable ground-truth function [14]. Empirical evidence suggests that this issue persists across state-of-the-art models, irrespective of increases in parameter scale or training data refinement. Consequently, the inherent fallibility of LLMs necessitates the deployment of robust mitigative frameworks, such as RAG or external knowledge integration, to ensure operational safety in high-stakes applications [14].

The adaptation of LLMs to domain-specific tasks remains challenging due to difficulties in parsing technical terminology and maintaining conversational context within complex environments, such as telecommunications product specifications [15]. These limitations stem from the general-purpose nature of the underlying training data, which often lacks the specialized depth necessary for industrial applications. Consequently, models deployed in high-stakes sectors like healthcare may generate vague or misleading outputs that fail to align with established clinical guidelines [16], necessitating robust fine-tuning or the integration of external knowledge bases.

## C. Approaches to Address LLM Limitations

The domain specialization of LLMs is increasingly recognized as a strategy to overcome their limitations in handling nuanced, domain-specific tasks. The comprehensive survey by Reference [11] emphasizes that domain specialization can address key issues like hallucination, biases, and the inability to generalize to specific contexts. The paper introduces a systematic taxonomy of techniques to tailor LLMs for specialized applications, categorized into black-box, grey-box, and white-box approaches.

Black-box methods involve external augmentation, where LLMs interact with domain-specific knowledge repositories or tools without modifying internal model parameters. For instance, retrieval-based augmentation integrates real-time, task-specific information from external databases, enhancing the model's accuracy without retraining. Grey-box methods focus on prompt crafting, including designing discrete (zero-shot or few-shot) and continuous prompts to steer LLM responses. White-box methods allow full access to model internals for fine-tuning or parameter updates. Techniques like adapter-based fine-tuning or instruction-tuning refine LLMs to align closely with domain-specific objectives [17].

Computational and ethical challenges arise when deploying domain-specialized LLMs, including ensuring up-to-date knowledge, maintaining transparency, and mitigating biases. Recent research has highlighted two key strategies for integrating domain-specific knowledge into generative models: RAG and fine-tuning. RAG supplements model inputs with external data dynamically, while fine-tuning incorporates domain-specific information directly into the model's parameters [17].

The research analyzed these methods in the context of agricultural applications, utilizing a pipeline that included data acquisition, Q&A generation, and optimization. The research demonstrated that fine-tuning led to a 6% improvement in model accuracy, which increased further by 5% when combined with RAG. Fine-tuning proved particularly effective in processing geographically specific data, enhancing response similarity and improving the model's ability to generate precise, succinct answers. However, the process requires substantial computational resources and careful data preparation. In contrast, RAG dynamically accesses external knowledge without altering the underlying model, offering a flexible approach for handling scenarios with frequent updates to the information base [18].

## D. AI Virtual Assistants in Water Management

The integration of AI and Big Data Analytics into WRM has emerged as a transformative approach to address the increasing complexity of managing water systems. Traditional WRM methods often struggle to meet the demands of real-time data acquisition, effective analysis, and intelligent decision making. AI technologies, such as machine learning and deep learning, enable precise water quality monitoring, allocation, and demand forecasting by leveraging diverse data sources. Big Data Analytics further complements these applications by integrating and analyzing large datasets, providing actionable insights, and facilitating real-time monitoring of water availability and quality. Moreover, participatory decision making is enhanced by AI and Big Data Analytics, as they foster transparency and collaboration among stakeholders. These technologies ensure that WRM practices are sustainable and resilient, marking a paradigm shift toward adaptive and effective resource management strategies.

Recent advancements in AI, particularly LLMs, offer scalable solutions to complex environmental challenges. A notable application is the BlueGAP AI Hub, which utilizes a suite of conversational agents to manage water quality and nitrogen pollution [19]. The framework consists of specialized agents—including a Virtual Champion, Nitrogen Expert, Local WQ Expert, Action Planner, and Data Expert—each optimized for distinct facets of water management. By synthesizing diverse data sources such as local insights, scientific reports, and government surveys, these agents provide region-specific strategic planning and statistical analysis. Technical precision is maintained through the integration of NER, advanced data retrieval, and context-aware dialogue systems, resulting in an overall accuracy rate exceeding 89% [19]. This architecture demonstrates the efficacy of conversational AI in translating complex scientific knowledge into actionable interactions, establishing a benchmark for scalable platforms in sustainable environmental conservation [19].

Further specialization is evidenced by WaterGPT, a bilingual LLM built on the Qwen2-2B-Instruct model and designed for hydrology applications [20]. The research addresses challenges in applying AI to hydrology, focusing on data analysis for water resources (text and images), intelligent decision making, and interdisciplinary integration through question-answering systems. WaterGPT employs a structured training methodology, combining incremental pretraining and supervised fine-tuning (SFT) with a specialized dataset from hydrology dictionaries, encyclopedias, and domain-specific literature, annotated into knowledge-based, task-oriented, and multi-turn dialogues. Additionally, the multi-agent framework, Water_Agent, executes complex hydrological tasks modularly, incorporating multimodal data processing and optimized with LoRA and DoRA for memory-efficient fine-tuning, balancing domain expertise with generalization. WaterGPT was evaluated using the EvalWater dataset, comprising over 10,000 questions across 21 hydrology subdomains, achieving an accuracy of 33.04%, outperforming GPT-4 by 12.33 percentage points. However, in the general Ceval dataset, WaterGPT's performance was eight points below GPT-4, indicating a trade-off between specialization and generalization. The Water_Agent framework demonstrated robust task execution, achieving a 43% task completion rate across both simple and complex scenarios, underscoring WaterGPT's potential in addressing domain-specific challenges in hydrology [20].

The deployment of Waterbot in Arizona serves as a primary case study for utilizing RAG to deliver context-sensitive information within water-scarce regions [21]. Waterbot's unique contributions lie in its interdisciplinary approach, incorporating technical, cultural, and user-centered considerations. Notably, the integration of Indigenous perspectives provides a holistic understanding of Arizona's water issues. Using publicly available information from the 22 tribal nations, the chatbot represents historical, ecological, and spiritual relationships with water, alongside municipal and agricultural perspectives. This integration addresses longstanding biases in hydrological data and supports equitable decision making. Initial feedback from urban, rural, and tribal communities indicates high user satisfaction when using Waterbot, though challenges remain in ensuring up-to-date data and continuous engagement with diverse stakeholders. The paper underscores the advantages of custom-built chatbots, particularly in public-focused applications. Unlike generic GPT-based systems, Waterbot's architecture allows for fine-grained control of data, user interaction feedback, and iterative updates [21].

Researchers in hydrology and environmental sciences encounter significant barriers due to the complexity of hydrological models, rapid iterations of software frameworks, and extensive technical documentation [22]. These tasks necessitate a specialized intersection of domain expertise and advanced programming proficiency. To mitigate these complexities, HydroSuite-AI provides an AI-powered assistant designed to streamline hydrological research and data analytics [22]. The platform integrates advanced LLMs with open-source hydrology-related libraries, facilitating the generation of context-aware code snippets and the automated retrieval of hydrological data. By offering a web-based interface for resolving factual queries and performing data analysis, HydroSuite-AI enables the seamless incorporation of AI into existing WRM and modeling workflows, thereby reducing the operational overhead associated with modern hydrological modeling [22].

The technical landscape of AI in water management has moved beyond general chatbots toward specialized, problem-solving systems. A significant advancement in 2025 is the IWMS-LLM (Intelligent Water Resources Management System), which utilizes LLMs as a core engine for visual workflow orchestration [23]. Developed to address the complexity of multi-stage hydrological tasks, IWMS-LLM employs a 'problem-driven service model' where the LLM interprets natural language requests to automatically invoke domain-specific models, such as flood propagation simulations or station-level data retrieval. This architecture effectively bridges the gap between human users and technical computational models, democratizing access to complex water management tools [23].

For the specific challenge of urban flood resilience and community communication, the FLAI (Flood Large Language Model AI) framework has been introduced [24]. FLAI integrates Retrieval-Augmented Generation (RAG) with a specialized Flood Knowledge Graph that aggregates data from social media, meteorological sensors, and historical flood reports. Its primary contribution is the generation of contextualized narratives, maps, and infographics to reduce the cognitive burden on decision-makers during extreme weather events. Comparative studies indicate that the FLAI system reduces factual inconsistency in flood-related decision support by over 75% compared to general-purpose LLMs [24].

Expanding upon the requirements for domain-specific, high-fidelity response frameworks, the ACQUIRE AI Assistant has been developed as an AI-augmented system for drinking water quality event response [25]. The system addresses the sector's "data-rich but information-poor" dilemma by employing a RAG pipeline to mine and structure knowledge from years of unstructured regulatory reports, specifically those published by the Drinking Water Inspectorate (DWI). This approach facilitates a shift from reactive to proactive modeling by providing water professionals with a conversational interface capable of synthesizing historical incident data to generate real-time response scenarios and structured action plans for contamination events. Operational testing within a utility control room environment confirmed that the platform serves as an effective "roadmap" for inexperienced staff while functioning as an augmentation tool for experts to validate initial response strategies. By transforming siloed documentation into a dynamic, auditable knowledge base, this framework enhances the safety, regulatory compliance, and resilience of drinking water supplies [25].

Finally, the development of HydroLLM represents a foundational step toward creating a "hydrology-native" artificial intelligence [26]. The study systematically fine-tuned several large language models using a specialized corpus of ~8,800 hydrology question–answer pairs drawn from textbooks and recent research articles [26]. Results demonstrated that the 8B-parameter DeepSeek-Llama variant achieved the strongest overall performance across multiple instructional formats, outperforming both larger models (such as the 32B variant, which overfit severely) and smaller models (which underperformed) [26]. This underscores that larger size is not always advantageous and highlights the critical importance of matching model capacity to dataset size to ensure technical precision and robust generalization in hydrological reasoning and decision support [26].

To situate our contribution within the broader landscape of AI applications in water management, Table 1 presents a summary of prominent LLM-powered systems developed for water-related use cases across different regions.

TABLE I  EXISTING LLM-BASED SYSTEM INITIATIVES IN THE WATER SECTOR

| Chatbot Name / Platform | Description | Country / Region | Reference |
|---|---|---|---|
| Blue GAP AI Hub | A suite of LLM-powered conversational agents addressing water quality issues such as nitrogen pollution, offering localized, actionable advice. | United States | [19] |
| WaterGPT | A bilingual LLM developed for hydrology applications with multi-agent task execution and domain-specific training. | China | [20] |
| Waterbot | A RAG-based chatbot offering water-related information to Arizona | United States | [21] |

| | residents, integrating Indigenous knowledge and public data. | | |
|---|---|---|---|
| HydroSuite-AI | AI-powered assistant that supports hydrological research by generating code, analyzing data, and providing context-aware help. | United States | [22] |
| IWMS-LLM | An intelligent system using LLMs as a central orchestrator to bridge natural language queries with domain-specific models like EPANET. | China | [23] |
| FLAI | A flood-risk communication platform using Knowledge Graphs and RAG to produce contextualized narratives and infographics for decision support. | United States | [24] |
| ACQUIRE AI Assistant | A RAG-based platform that mines historical regulatory reports to generate real-time response plans and decision support for water quality incidents.. | United Kingdom | [25] |
| HydroLLM | A domain-specific LLM fine-tuned on a 8,800-pair hydrology dataset, designed for high-fidelity technical reasoning and research assistance. | United States | [26] |

### III. METHODOLOGY

This research employs advanced AI methodologies to design and develop WaterCopilot. A critical method of this approach is the LLM's tool-calling mechanism, which allows the chatbot to dynamically interact with predefined tools, execute specific tasks, and return accurate, context-aware responses.

#### A. Study Area and Data Collection

The data collection process for WaterCopilot was designed to address two critical needs in LRB management: access to authoritative policy documents and real-time monitoring of hydrological conditions. Data was categorized into two primary types to ensure comprehensive coverage.

*1) Static Documents*

Static documents form the foundational layer of WaterCopilot's knowledge base. These documents contain structured, policy-driven, and historically validated information essential for water governance, environmental monitoring, and hydrological modeling in the LRB. Key sources include Limpopo water management reports, transboundary policy frameworks, hydrological model documentation, and environmental flow assessment studies. These documents provide essential background on rainfall patterns, river flow dynamics, environmental thresholds, and basin-wide resource management practices. To facilitate efficient retrieval, these static documents have been processed, indexed, and embedded using Azure AI Search, enabling semantic search capabilities within the WaterCopilot system.

*2) Real-Time Data*

Real-time data forms a critical component of WaterCopilot's functionality, enabling the system to deliver current, historical, and predictive insights for water resource management across the Limpopo River Basin. This dynamic data is retrieved from several live databases and APIs maintained by the IWMI and partner monitoring networks. The integrated real-time data streams include:

- **Water Resources Monitoring:** Access and analyze real-time, historical, and forecasted data on rainfall, river discharge, and reservoir levels from basin-wide monitoring stations and modeling systems.
- **Environmental Flows (E-flows):** Screen for e-flow risks to protect river ecosystem health using calculated thresholds, alerts, and trend analyses derived from both historical records and forecasted hydrological conditions.
- **Water Accounting:** Obtain comprehensive snapshots of the basin's water balance by integrating data on water availability, allocations, use, and fluxes across sectors.
- **Irrigated Areas and Agricultural Water Use:** Identify and estimate agricultural water consumption for specific fields using satellite-derived data and irrigation records, supporting water-use efficiency and allocation planning.

Data streams from these sources are accessed via dedicated APIs and updated CSV files, ensuring that WaterCopilot remains responsive to evolving environmental conditions. By integrating this suite of real-time, historical, and forecasted hydrological, meteorological, and water-use data, the system supports informed decision-making, timely environmental alerts, and proactive water management strategies.

#### B. High-Level Architecture

The proposed system architecture comprises four key components as illustrated in Fig. 1:

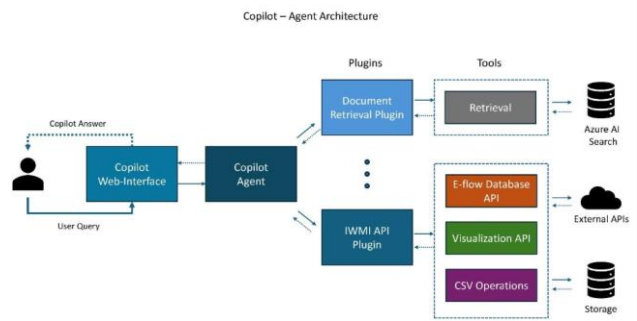

*Figure 1: High-Level Overview of the WaterCopilot Agent Architecture and Its Integration with Plugins and Tools*

- **WaterCopilot Web-Interface:** The user interface provides a seamless communication layer between users and the WaterCopilot system. It captures natural language queries, forwards them to the WaterCopilot Agent for processing, and presents structured, context-aware responses. Designed for clarity and ease of use, the interface ensures efficient access to critical water resource information for a diverse range of users.

- **WaterCopilot Agent**: The Orchestration Layer is a core component of the system that manages the flow of tool invocation of the LLM. It is responsible for gathering and organizing the necessary details of available tools and plugins, ensuring that the LLM can effectively call the relevant tools in response to specific queries. This layer ensures seamless coordination and utilization of tools, enabling the LLM to execute tasks efficiently while maintaining optimal query handling and tool usage.
- **Plugin and Tools:** The Document Retrieval Plugin (*iwmi-doc-plugin*) enables the WaterCopilot Agent to query and retrieve static documents through Azure AI Search. When a query requires document-based information, the plugin fetches the relevant content from the indexed data. The IWMI API Plugin (*iwmi-api-plugin*) facilitates access to dynamic data by interacting with external APIs and CSV-based sources, allowing the WaterCopilot Agent to retrieve real-time information such as environmental flow data and generate visualizations.
- **External Resources:** Azure AI Search is a cloud-based service utilized for processing, indexing, and searching static documents. It offers robust search capabilities that are leveraged by the Document Retrieval Plugin. In addition to this, external APIs are integrated to interact with databases and CSV-based data sources, providing access to dynamic information such as real-time environmental flow data.

C. Detailed Implementation

1) Web Interface

This interface directly interacts with the user, serving as the primary component for communication. It enables users to input natural language queries and receive responses in a seamless conversational format. The interface accepts the query and forwards it to the Copilot Agent for processing. Once the Copilot Agent retrieves the necessary information from the relevant plugin—either static documents or real-time data APIs—the response is displayed back to the user in a clear and structured manner.

When a new user opens the page, the interface presents a few predefined options to help them explore what the chatbot can do. These options make it easier to get started and understand the available features. Users can choose the **Limpopo Library** to ask questions about static documents such as hydrological model reports, policy papers, water governance frameworks, and environmental assessments relevant to the Limpopo River Basin. The **Limpopo Real-Time Analysis** option allows users to access dynamic environmental data, including rainfall patterns, river flow rates, and environmental flow thresholds from live monitoring stations. If they need to retrieve or save structured results, the **Export/Generate Data** option allows users to extract summarized responses, tabular outputs, and source references in downloadable format. There is also a **New Conversation** option for those who prefer to start fresh without selecting a predefined topic. This design ensures a smooth and intuitive experience, making it easy for users to navigate the chatbot and find the information they need for effective water resource management.

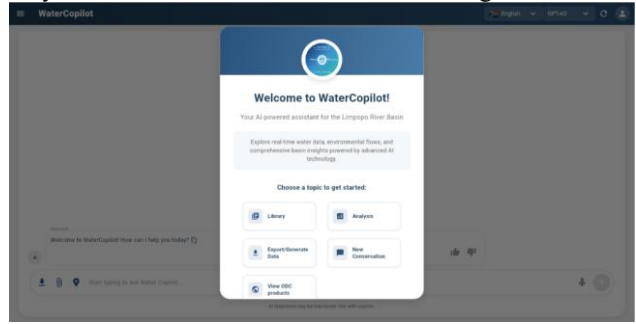

*Figure 2: Overview of initial user options in WaterCopilot*

2) WaterCopilot Agent

The WaterCopilot Agent serves as the core logic engine of the system. It manages the flow of LLM requests and determines how tools and plugins are made available for effective query handling. Upon receiving a user query from the Chatbot Web Interface, the agent gathers all the available tools along with their details such as descriptions and parameters and provides this information to LLM. The LLM evaluates the query and decides whether to call an external tool. If the LLM determines that a tool invocation is required, it invokes the appropriate tool, retrieves the response, and interprets the result for the users.

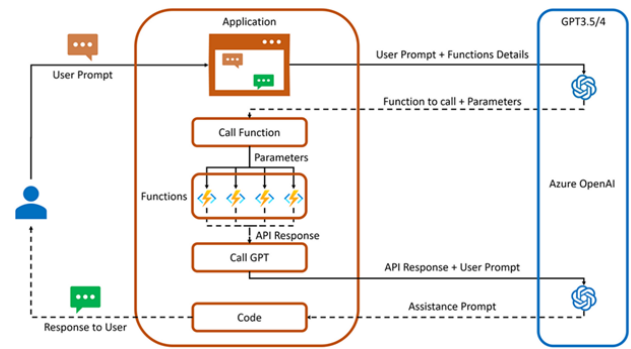

*Figure 3: Detailed architecture of the tool calling mechanism in agent*

The following steps are involved in tool calling mechanism:

- **User Prompting:** Users interact with the chatbot by providing the queries which act as an input for the chatbot.
- **Tool Meta-data Retrieval:** Each time the user submits a query, the agent calls a specific endpoint to retrieve a list of available tools along with their descriptions. This retrieval includes detailed metadata for each tool outlining its capabilities.
- **Determining the Tool:** Based on the user's input, the LLM evaluates whether the request can be handled using its built-in language processing capabilities or whether it requires the assistance of an external tool or function. The LLM compares the user's input against the retrieved tool descriptions to determine if a tool is necessary.
- **Invoking Tools:** If the LLM identifies that a tool is required, it proceeds to call the relevant tool with

the necessary arguments (parameters) extracted from the user's input. The LLM communicates directly with the tool, sending any data required for the tool's operation.

- **Receiving and Processing Tool Responses:** Once the tool has been executed, the LLM receives the response generated by the tool. The LLM then processes and interprets this data and gives the response back to the user.

*D. Tool-Calling Mechanism*

The tool-calling mechanism enables dynamic interactions between LLMs and external knowledge sources. Each tool is annotated with descriptions, defined arguments, and designed for specific actions. The mechanism involves: user prompting, tool metadata retrieval, tool determination, tool invocation, and response processing.
Plugin is a collection of tools which enhance modularity and scalability. Each plugin encompasses tools related to a particular category of operations, such as data retrieval, computation, or external API interactions. This organizational structure simplifies the management of tools and supports the efficient expansion of functionalities within the chatbot system.

*E. Development of Plugin and Tools*

Two custom plugins, which are **IWMI-DOC-PLUGIN** and **IWMI-API-PLUGIN**, have been developed to support WaterCopilot's ability to handle both static document retrieval and real-time data analysis. Each plugin is designed as a standalone service and communicates with the Copilot Agent through HTTP endpoints, enabling seamless integration and independent operation.

*1) IWMI-DOC-PLUGIN*

The **Document Retrieval Tool** under the **IWMI-DOC-PLUGIN** is responsible for handling semantic search over static documents such as policy reports, hydrological models, and environmental flow assessments related to the Limpopo River Basin. These documents are pre-processed, chunked, and embedded into vector representations, which are then indexed using Azure AI Search. When a user submits a query related to static content, the plugin performs a similarity search against this index and returns the most relevant passages for response generation.

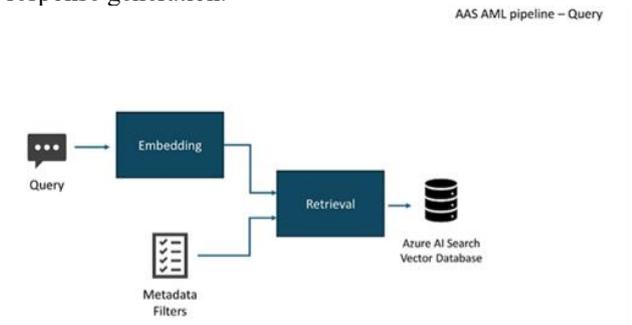

*Figure 4: Workflow of Copilot Querying Static Documents with Embedding and Retrieval Processes*

The following steps are involved in query processing from azure index:

- **User Input:** The process initiates when a user submits a query through the Chatbot interface.
- **LLM understands the query and tool invocation:** The LLM processes the submitted query and invokes the Document Retrieval Tool with the appropriate parameters.
- **Embedding Process:** The submitted query undergoes an embedding process, where it is transformed into a vector representation. This transformation utilizes the same embedding model applied to the indexed documents, ensuring that both the query and documents reside in the same semantic space. This alignment is critical for achieving accurate similarity matching.
- **Metadata Filtering:** Following the embedding process, the query vector representation is combined with metadata filters proposed by the agent LLM. These filters refine the search results by applying specific criteria, such as document type and relevance, to narrow down the scope of the retrieved information. This step ensures that only the most applicable static documents are considered in the retrieval process.
- **Semantic Search:** After applying filters, the tool leverages query embeddings and metadata to search within the Azure Index. It then retrieves the most relevant snippets from static documents that closely match the user's query. This process ensures that the chatbot delivers contextually relevant results, enhancing response accuracy and aligning with the user's intent.

*2) IWMI-API-PLUGIN*

The tools under the **IWMI-API-PLUGIN** are designed to dynamically retrieve and process real-time environmental data related to the Limpopo River Basin. These tools interact with external APIs connected to institutional databases, enabling the chatbot to access updated information on rainfall patterns, river flow levels, and environmental flow alerts from multiple monitoring stations.

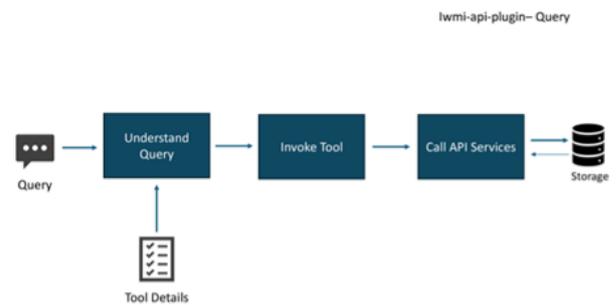

*Figure 5: Workflow of Copilot Querying Databases via API Calls Using the IWMI API Plugin*

The following steps are involved in query processing:

- **Query Submission:** The process begins when a user submits a natural language query through the chatbot interface.
- **Query Understanding and Parameter Extraction:** The Agent fetches the tool details and analyzes the user query to determine its intent and whether access to real-time environmental data is required. If applicable, it extracts key parameters from the query, such as the type of data requested (e.g., rainfall, river flow, or environmental flow alerts), specific monitoring stations, and the relevant time range.

- **Invoke Tool:** With the extracted parameters, Agent calls the tool, setting up the API request with the specific parameters required. The tool then executes the API call.
- **Content Processing and Presentation:** After the API call is made, Agent fetches the relevant information from multiple databases, including IWMI-DB, INWARDS-DB, and FISHTRAC-DB. This step gathers data from various sources, compiling the information needed to respond to the user's query.

*F. Indexing Pipeline*

The selected documents, including policy reports, hydrological models, and environmental flow assessments, are processed through a specialized pipeline to create an Azure AI Search index. This indexing enables efficient retrieval of static information, allowing the Document Retrieval Tool to query the data effectively and provide relevant, context-aware responses to water-related inquiries.

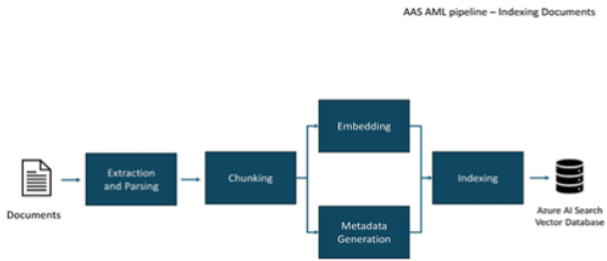

*Figure 6: Pipeline Architecture for Generating an Azure Index from Limpopo PDF Documents*

The index generation has the following steps:

- **Documents:** The pipeline begins with source documents, such as PDF files containing static information about the Limpopo basin. These documents could include research reports, policy documents, or environmental data.
- **Extraction and Parsing:** The process begins with extracting raw content from documents, including text, tables, and other structured elements, using Azure Document Intelligence services. This enables the efficient extraction of both structured and unstructured data. Once extracted, the content is parsed to identify its structure and categorize different types of information, such as separating and classifying paragraphs, headers, and tables based on their roles in the document.
- **Chunking:** Large documents are divided into smaller, manageable pieces or "chunks." This is crucial to ensure that each chunk is small enough for efficient processing, especially during the embedding and querying phases. Chunking improves the efficiency of the indexing process by handling content at a granular level, enabling more precise search results by reducing the document's complexity.
- **Embedding:** Each chunk is converted into a vector representation (embedding) using a pre-trained deep learning model. These embeddings capture the semantic meaning of the text, enabling more accurate search and retrieval. Pre-trained models are used to create high-dimensional vectors for each chunk, allowing efficient comparisons and search operations that go beyond keyword matching.
- **Metadata Generation:** Metadata is created for each chunk, which includes attributes such as the document's title, date, document type, and relevant tags. Metadata enhances the indexing process by providing additional context, improving the relevance of search results, and helping categorize the information for better retrieval.
- **Indexing:** The final step involves indexing the processed chunks and their embeddings into a vector-based search index, allowing efficient searching based on semantic meaning rather than exact keywords. Azure AI Search Vector Database is used for indexing, ensuring that the content is searchable based on its semantic meaning, along with the associated metadata for refined, context-aware search results.
- **Searchable Azure AI Index:** The documents are stored in a structured format that enables fast and accurate searches using Azure's AI-powered search capabilities. This index allows the Document Retriever Tool to efficiently process queries and return relevant results.

## IV. EVALUATION AND RESULTS

The core architecture of WaterCopilot is built around the integration of RAG and tool-calling capabilities, to provide users with precise and contextually relevant information in the field of water management. Evaluating the performance of such a system requires a comprehensive benchmarking strategy that assesses both the retrieval and generation components.

The RAGAS (Retrieval-Augmented Generation Assessment) framework offers a structured approach for this evaluation, enabling the measurement of key performance metrics without relying solely on human-annotated ground truth data [27]. By adopting this framework, the WaterCopilot chatbot's effectiveness in retrieving and generating information can be rigorously evaluated, ensuring its reliability and accuracy in facilitating informed water management decision-making.

*A. Evaluation Methodology*

*1) Evaluation Process*

To systematically evaluate the performance of the WaterCopilot, a structured process was designed to ensure a reliable assessment. The evaluation involved the following steps:

- **Question Selection:** A total of 30 questions were curated from two primary sources: indexed static documents and real-time environmental data. These questions were designed to comprehensively evaluate the system's full capabilities, ranging from document-based information retrieval to real-time data interpretation. The question set includes tasks requiring interpretation, comparison, and numerical calculations, thereby assessing performance across both simple factual queries and complex analytical scenarios. The questions were stratified into three difficulty levels: Level 1 (direct factual retrieval, questions 1-13), Level 2 (document-based analytical questions, questions 14-21), and Level 3 (complex real-time analysis, questions 22-30). The complete evaluation dataset is provided in Appendix B.
- **Ground Truth Compilation:** For each question, a ground truth was identified and validated. This serves as a benchmark for evaluating the chatbot's responses.
- **Evaluation Execution:** Each question was individually passed through the chatbot, during which the retrieved resources used by the chatbot to formulate its responses were recorded, and the generated answers were documented for further analysis.

2) *Evaluation Metrics*

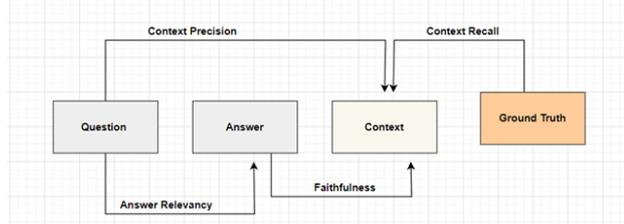

*Figure 7: RAG Evaluation Metrics and Relationship*

- **Context Precision:** Measures the proportion of retrieved context that is relevant to the given query. High precision ensures that the system retrieves predominantly pertinent information, minimizing irrelevant data [27].
- **Context Recall:** Evaluates how effectively the retrieved context encompasses the relevant information required to address a query. A high recall score signifies that the system has successfully captured most of the essential context necessary to answer the query comprehensively [27].
- **Answer Relevancy:** Evaluates how pertinent the response is to the original query, ensuring that the answer directly addresses the user's question. Higher relevancy signifies that the response aligns well with the query's intent [27].
- **Faithfulness:** Assesses whether the response is supported by the retrieved context, ensuring that the answer does not introduce information not grounded in the provided context [27].

3) *Evaluation Results*

To assess the chatbot's performance, the evaluation process was carried out with 30 questions with different difficulty levels such as low, medium and complex. The metrics such as faithfulness, answer relevancy, context precision and context recall were calculated for each question separately. From there the following mean values were obtained for the set of questions.

TABLE II. EVALUATION RESULTS

| Evaluation Metric | Mean Value |
|---|---|
| Faithfulness | 0.7877 |
| Answer Relevancy | 0.8571 |
| Context Precision | 0.8009 |
| Context Recall | 0.7763 |

The RAGAS framework validation architecture and the computational flow between these metrics are detailed in Appendix A (Fig. 12), illustrating how the individual metric scores combine to produce the overall system assessment.

  a) *Evaluation for the Retrieval Component*

The retrieval component plays a crucial role in the WaterCopilot chatbot, as it ensures that user queries are addressed by accurately extracting relevant and contextually appropriate information from indexed documents and APIs that query the database. This component forms the foundation for delivering precise responses, enabling users to access reliable water-related information seamlessly, particularly in the context of the Limpopo River Basin.

Among the discussed metrics, Context Precision and Context Recall are vital in measuring the effectiveness of the retrieval process. Context Precision, with a mean value of 0.8009, reflects the system's ability to retrieve predominantly relevant information, minimizing irrelevant data. Context Recall, with a mean value of 0.7763, highlights the system's capacity to comprehensively capture all relevant information required to answer a query. These scores demonstrate the chatbot's balanced performance in ensuring both relevance and coverage. Collectively, these metrics highlight the effective design and robust performance of the retrieval components, specifically the Document Retriever and API tools. The results underscore the capability of these tools in efficiently delivering relevant water-related information, thereby showcasing their effectiveness in facilitating comprehensive and accurate retrieval for decision-making in the Limpopo River Basin.

  b) *Evaluation for the Generation Component*

The generation component is responsible for synthesizing accurate, contextually aligned, and coherent responses derived from the retrieved information. Its effectiveness is assessed using metrics that evaluate the consistency of the generated answers with the provided context and their relevance to the user's query. The generation component is more critical than the retrieval component because it directly impacts the accuracy of the generated answer, which significantly influences user satisfaction. While the retrieval component focuses on sourcing relevant information, the generation component is responsible for synthesizing that information into a coherent and contextually aligned response. The quality of the generated answer determines how well it meets the user's expectations, affecting their satisfaction with the system. Therefore, the effectiveness of the generation component plays a vital role in ensuring both the accuracy of the response and the overall user experience.

The Faithfulness metric measures the factual consistency of the generated answer with the retrieved context. It ensures that all claims made in the response can be inferred directly from the context, preventing hallucinations or unsupported information. The mean Faithfulness score of 0.7877 indicates that the majority of the generated responses were consistent with the context provided.

The Answer Relevancy metric evaluates how well the generated response aligns with ground truth values. This metric reflects the chatbot's ability to produce relevant and meaningful answers that match the user's intent and the ground truth. A mean score of 0.8571 indicates the chatbot's strong capability in generating responses that are both pertinent and valuable to the user.

### c) Overall System Evaluation using RAGAS Score

To provide a unified assessment of the WaterCopilot chatbot's performance, the RAGAS Score was utilized. By combining these four key metrics, the RAGAS Score captures the effectiveness of both the retrieval and generation components in delivering high-quality outputs. The retrieval process ensures that the most relevant information is identified and used, while the generation component guarantees that the system produces responses that are not only accurate but also aligned with the user's intent. This holistic evaluation provides a nuanced understanding of the chatbot's overall performance in real-world tasks, accounting for both the precision and relevance of the retrieved context as well as the fidelity and coherence of the generated response.

The harmonic mean of these four aspects—Context Precision, Context Recall, Faithfulness, and Answer Relevancy—gives the RAGAS Score [27], which serves as a single measure of the performance of the chatbot system across all critical evaluation dimensions. This method ensures that lower performance in any metric disproportionately impacts the overall score, emphasizing the importance of balanced performance across all aspects.

For this study, the individual metric scores were as follows: Precision = 0.8009, Recall = 0.7763, Faithfulness = 0.7877, and Relevancy = 0.8571. Using the harmonic mean of these scores, the RAGAS Score was calculated to be approximately 0.8043.

The final RAGAS Score of 0.8043 reflects the overall performance of the chatbot in both retrieval and generation tasks. This score suggests strong overall performance, indicating that the system is effective in retrieving relevant context and generating accurate, contextually appropriate responses.

## V. DISCUSSION

### A. Key Findings

The WaterCopilot chatbot's evaluation demonstrates strong performance in both retrieval and generation tasks, addressing information needs related to water resource management. The Context Precision (0.8009) and Context Recall (0.7763) metrics highlight its ability to retrieve relevant and comprehensive context from indexed documents and databases. These metrics are fundamental to generating high-quality responses, as the accuracy and completeness of the retrieved context directly influence the chatbot's ability to provide relevant and contextually appropriate answers.

The retrieval component also enhances computing performance and cost efficiency. High precision reduces unnecessary data processing, minimizing token usage during response generation. As the cost of language model operations scales with token usage, efficient retrieval reduces computational expenses. Additionally, high recall ensures critical information is retrieved on the first attempt, eliminating the need for iterative queries or re-processing. Maintaining a balance between precision and recall allows the chatbot to deliver accurate responses while optimizing resource consumption. It is worth noting that a chatbot can sometimes generate highly relevant and faithful answers even with low retrieval precision and recall. This occurs when the language model selectively utilizes only the most relevant portions of the retrieved context, ignoring irrelevant details. While this capability highlights the robustness of the generation component, it exposes inefficiencies in the retrieval process. Excessive or irrelevant context increases computational costs and token usage unnecessarily, even if the final response appears accurate. Consistently achieving high retrieval precision and recall is therefore essential for ensuring both efficiency and long-term reliability.

The chatbot's generation capabilities are also noteworthy. The Faithfulness score (0.7877) demonstrates that most responses align with the retrieved context, minimizing hallucinations. However, the chatbot's reliance on its inherent knowledge during response generation can affect this metric. When additional information provided by the model is true but not found in the retrieved context, it is not accounted for in the Faithfulness score, potentially underestimating the model's alignment with facts. Conversely, false or unsupported claims from inherent knowledge directly lower the score. The Answer Relevancy score (0.8571) further highlights the chatbot's ability to generate meaningful and contextually relevant responses. This performance reflects the effective integration of retrieval-augmented generation, where retrieved context serves as the foundation, and the language model enhances clarity and coherence. Together, these results validate the chatbot's ability to deliver accurate, efficient, and user-focused responses in water resource management contexts.

### B. Key Features and Innovations

#### 1) Guided Interaction for Queries

One of the standout features of WaterCopilot is its ability to simplify complex queries through guided interaction. When users begin exploring a topic, Copilot prompts them step by step, ensuring that all necessary parameters are specified for an accurate response. This structured approach helps users provide relevant information, leading to more precise and effective results. For example, if a user inquiries about rainfall data, the bot might ask for a location, time period, and any additional details that could refine the search. If users are unsure about what inputs are needed, the Copilot provides helpful examples and suggestions, making the process smoother and less intimidating. This structured approach is especially useful for multi-part queries, where the bot walks users through each step, ensuring that no critical details are overlooked. By guiding users through the process, WaterCopilot helps them formulate effective questions, leading to more precise and relevant answers, thereby saving time and enhancing their understanding of the information they seek.

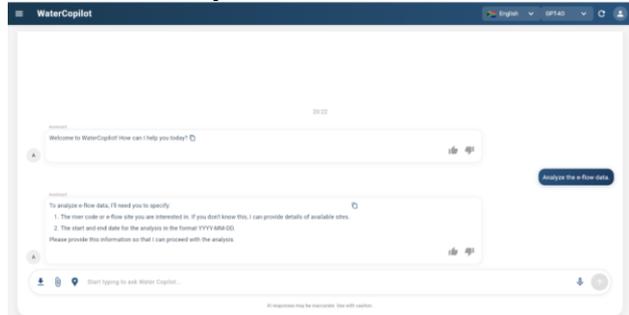

*Figure 8: Guide on how the WaterCopilot assists users by providing detailed parameter information*

### 2) Reference Provision

Transparency and reliability are at the core of WaterCopilot's design. For every answer provided, WaterCopilot includes detailed references to the sources used. This ensures that users can verify the information and trust its accuracy, allowing them to trace back to the original data and sources for further validation. When information is retrieved from static documents, the bot specifies which document was referenced, highlighting the exact snapshots or sections from which the data was drawn. Similarly, for real-time data fetched from APIs, the bot provides real-time references, ensuring that the information reflects the latest updates. This level of transparency not only builds trust but also facilitates further research by enabling users to trace back to the original source material. Users can rely on these detailed references to ensure the credibility of the information presented, making WaterCopilot a dependable tool for research and decision-making.

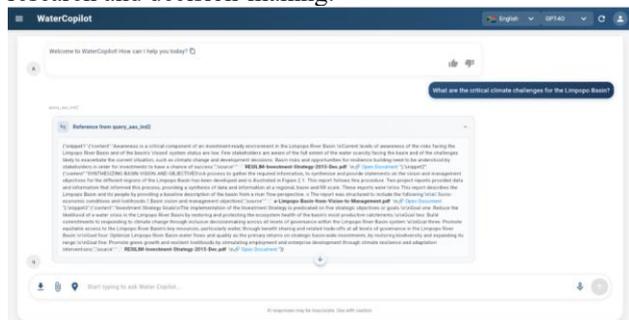

*Figure 9: Source references are provided by the Copilot when answering user queries*

### 3) Multilingual Support

Recognizing the diverse linguistic backgrounds of its users, WaterCopilot offers robust multilingual support. Users can interact with the bot in multiple languages, including English, French, Portuguese, and others, ensuring that language is not a barrier to accessing vital information. Whether a user asks a question in English or another supported language, the bot can process the query accurately and provide responses in the same language, ensuring a smooth and accessible experience. This feature is particularly important in regions like the Limpopo Basin, where stakeholders and users may speak different languages. By enabling communication across multiple languages, WaterCopilot broadens its reach and inclusivity, making it a valuable tool for a wider audience.

### 4) Summary and Insights

WaterCopilot is equipped with the capability to generate concise summaries and insights, making it easier for users to digest complex information. When users ask a question, the bot doesn't just provide raw data; it processes and combines relevant information from various sources to understand the overall context. By analyzing the data, the bot extracts key points and presents them in a clear, simplified summary, making complex topics easier to grasp. Additionally, the bot can identify patterns and highlight important details that might otherwise be overlooked, providing deeper insights into the subject matter. This feature ensures that users receive a comprehensive yet easy-to-understand response, enabling them to make informed decisions without being overwhelmed by too much data.

### 5) Calculation Capability

WaterCopilot is designed to handle a variety of calculations, adding significant value for users involved in water management and environmental assessments. Whether users need to calculate the average rainfall over a specific period, compare precipitation or e-flow values across different regions, or sort data for in-depth analysis, the bot can efficiently perform these tasks. This capability provides accurate and quick metrics, supporting data-driven decision making. By seamlessly integrating these computations into its responses, WaterCopilot helps users analyze data effectively, identify trends, and gain deeper insights, all within a single, interactive platform. This makes it a powerful tool for users who need precise data analysis.

*Figure 10: WaterCopilot's ability to perform calculations, including basic arithmetic and complex data computations*

### 6) Graphical Representations

To make data more accessible and easier to interpret, WaterCopilot offers advanced visualization capabilities. As shown in Figure 10, WaterCopilot can generate tailored charts and graphs by interacting with APIs that access data from the Limpopo Basin. These visual representations help users see patterns, trends, and comparisons at a glance, making complex information more digestible. Whether a user needs to visualize rainfall patterns over a period, analyze e-flow statistics, or compare water levels across different sites, the bot provides relevant graphs that simplify the information. By offering these visualizations, WaterCopilot helps users quickly grasp critical insights and make informed decisions. This feature is especially valuable for those who prefer visual data analysis over text-based data interpretation.

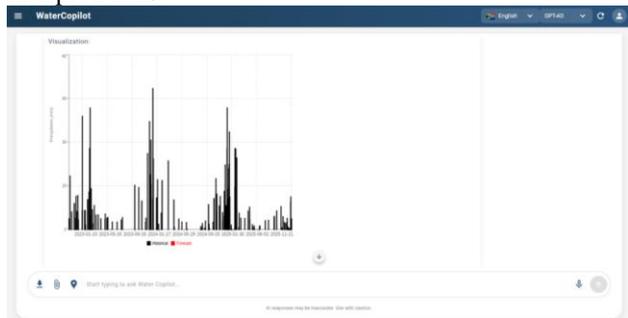

*Figure 11: Copilot's ability to generate visual representations of data*

WaterCopilot offers a range of powerful features designed to enhance user interaction and provide comprehensive information about the Limpopo River Basin. These features are documented in detail in the WaterCopilot User Guide [28], which provides step-by-step guidance for end-users.

### C. Limitations and Technical Challenges

This study acknowledges several limitations that have influenced the development and deployment of the developed WaterCopilot chatbot system:

- **Document Indexing Pipeline:** The pipeline employed for indexing static documents exhibits notable constraints. Specifically, the current implementation is unable to extract content from tables and images within documents. This limitation restricts the chatbot's ability to provide comprehensive responses when user queries relate to visual or tabular data.

- **Language Model Dependence:** The chatbot relies on GPT-based LLM as its core architecture, utilizing an external tool-calling mechanism to process user queries. While effective for the intended scope, this reliance limits compatibility with other advanced models, such as LLaMA. Adopting alternative models would require a substantial redesign of the system's core logic and implementation strategy, presenting a significant barrier to model flexibility and extensibility.

- **Multilingual Support Accuracy:** Although the chatbot is designed to understand and respond in multiple languages, this capability relies entirely on the GPT model being used. Consequently, the accuracy of multilingual responses may vary, and the system may occasionally fail to provide precise or contextually appropriate answers, particularly for complex or nuanced queries.

- **Evaluation Limitations:** The evaluation of the developed chatbot primarily relied on the RAGAS framework, which is grounded in metrics generated through LLM-based evaluations. While this approach provides valuable insights into aspects such as faithfulness, answer relevancy, context precision, and context recall, it lacks the involvement of extended human evaluation. The absence of human evaluative input, such as user satisfaction surveys or qualitative assessments, limits the understanding of the chatbot's real-world usability and effectiveness from a user-centric perspective. Incorporating human feedback, especially through measures of user satisfaction and interaction quality, could complement the LLM-based evaluation and provide a holistic view of the system's performance.

## VI. CONCLUSION AND FUTURE WORK

### A. Conclusion

This research focused on developing **WaterCopilot**, an AI-powered virtual assistant designed to support diverse stakeholders engaged in water resource management within the Limpopo River Basin. The system addresses the critical need for accessible, real-time, and context-aware information by integrating LLMs with advanced **RAG** and **function-calling** architecture tailored to the hydrological domain.

Through the combination of static document retrieval and live environmental data access, WaterCopilot demonstrates its ability to deliver accurate, timely, and relevant responses to queries related to rainfall, river flows, water availability, and policy documentation. Evaluation metrics, including performance across static and real-time data queries—highlight the system's robustness in both information retrieval and response generation. Key components such as the **Document Retrieval Tool** and the **Real-Time Query Tool** ensure high accuracy, while the system's capacity for contextual grounding supports factually consistent, trustworthy outputs.

In addition, WaterCopilot's conversational interface and topic-guided navigation enhance usability, enabling researchers, decision-makers, and basin authorities to efficiently access critical water-related insights. However, the system currently faces limitations in areas such as broader multilingual support, integration with multimodal datasets, and dependency on GPT-based LLMs. These constraints present opportunities for future enhancements, including improved data pre-processing pipelines, multilingual document indexing, and the incorporation of user-centered evaluation frameworks to strengthen operational relevance.

Overall, this study presents a strong foundation for the application of AI in water resource governance. WaterCopilot represents a scalable, adaptable solution that bridges data accessibility gaps and supports informed decision-making in transboundary river basins. By addressing its current limitations and incorporating advanced capabilities, the system holds promise as a transformative tool for sustainable water management and environmental resilience in Southern Africa.

*B. Future Directions*

As WaterCopilot continues to evolve, several key enhancements are planned to improve its functionality, user experience, and overall performance. The following outlines the primary areas of focus for future development:

- **Expansion of Static Data Sources:** One of the primary goals is to expand the repository of static data related to the LRB. By adding more documents and datasets, WaterCopilot will be equipped with comprehensive knowledge about historical patterns, environmental reports, and other relevant static information. This enhancement will provide users with a richer context for their queries, improving the overall depth of insights available through the bot.
- **Integration of Additional APIs:** To enhance the dynamic information available to users, future work will include the integration of more APIs. This will allow WaterCopilot to access a broader range of data sources, providing users with real-time insights into various aspects of the LRB. By expanding the range of data and services accessible through the bot, users will benefit from more accurate and timely information, facilitating better decision-making.
- **Support for Diverse Graph Types:** The capability to support various types of graphs and visualizations will be expanded. By implementing additional visualization techniques, users will be able to choose from different chart types that best represent the data they are analyzing. This flexibility will enhance the interpretability of the data and provide users with better insights into trends and patterns over time.
- **Access to the Digital Twin Main Portal:** In the future, WaterCopilot will implement a technique to enable direct access to the main Digital Twin portal from within the Copilot. This integration will allow users to seamlessly transition between the Copilot and the Digital Twin platform, providing a more cohesive experience. Users will be able to explore detailed data and features on the portal without needing to navigate away from the Copilot interface, further enhancing usability and accessibility.

## VII. ACKNOWLEDGEMENT

This study was funded by the CGIAR Initiative on Digital Innovation, which advances sustainable agrifood systems through digital solutions and the International Water Management Institute's Digital Innovations for Water Secure Africa (DIWASA) project. We also wish to thank all funders who supported this research through their contributions to the CGIAR Trust Fund (https://www.cgiar.org/funders/) and the Leona M. and Harry B. Helmsley Charitable Trust for their financial support for the DIWASA project.

We extend our sincere thanks to the Microsoft Research team for their technical collaboration in developing WaterCopilot. We are also grateful to Zolo Kiala and Surajit Ghosh for their insightful reviews and constructive feedback, which significantly enhanced the quality and clarity of this manuscript. Finally, we thank the Limpopo Watercourse Commission (LIMCOM) for their continued support throughout this project.

# APPENDIX

*A. Ragas Validation Overview*

The quantitative performance of the WaterCopilot was validated using the Ragas framework [27]. As illustrated in Fig. 11, the evaluation provides a comprehensive overview of the system's scoring across the previously defined metrics. The strong alignment observed between the retrieved hydrological context and the generated technical responses underscores the system's reliability and its capacity for deployment in high-stakes water management scenarios.

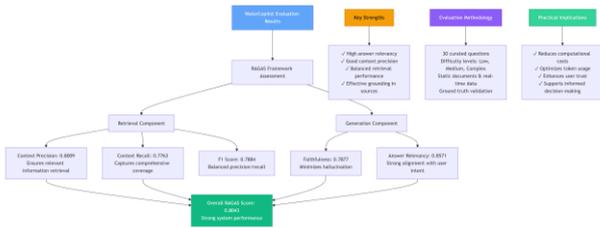

*Figure 12: Summary of Ragas validation metrics for WaterCopilot, demonstrating the quantitative alignment between retrieved context and generated responses*

*B. Evaluation Dataset*

*B.1 Document-Based Questions*

These questions test the system's ability to retrieve and present information from indexed static documents.

*Level 1: Direct Factual Retrieval*

1. What are the countries in the Limpopo Basin?
2. How many reservoirs are there in the Limpopo Basin?
3. What is the total water storage volume of all the dams?
4. What is the population of the Limpopo Basin?
5. What are the major economic activities in the Limpopo Basin?
6. What are the major rivers contributing to the Limpopo River flow?
7. What is the Mean Annual Runoff for the Limpopo Basin?
8. What are the major pollutants affecting water quality in the Limpopo Basin?
9. What are the key conservation areas in the Limpopo Basin?
10. What are the two identified transboundary aquifers within the Limpopo River Basin?
11. What infrastructure has the most significant water usage in the Limpopo River Basin?
12. What are the main water quality concerns in the Crocodile River?
13. What is the role of LIMIS in the basin's water management?

*Level 2: Document-Based Analytical Questions*

14. What are the final risk regions in the Limpopo River Basin?
15. What is the classification of water quality at selected sites based on the survey?
16. What is the impact of climate change on the Limpopo Basin?
17. What methods were used to determine the hydraulics for 21 sites across the Limpopo Basin?
18. What are the drivers of ecosystem change identified in the document?
19. What is the relationship between groundwater and surface water in the Limpopo River Basin?
20. Why is the upper Olifants River described as one of the most polluted rivers in southern Africa?
21. How do El Niño and La Niña events impact weather conditions in southern Africa?

*B.2 API/Real-Time Questions*

These questions test the system's ability to process dynamic data, perform calculations, and generate analytical insights from real-time sources.

*Level 3: Complex Real-Time Analysis*

22. Can you provide a list of all the available Eflow sites in the Limpopo Basin?
23. List the nearest rainfall stations to the Olifants River.
24. List the five nearest rainfall stations to the Luvuvhu River
25. Can you provide the monthly average rainfall analysis for Shisakashanghondo for the year 2024?
26. Can you provide a comparison of the monthly average rainfall analysis for Shisakashanghondo for the years 2024 and 2023?
27. What is the water availability for the Elephants River in the Limpopo Basin for March 2024?
28. Identify which river had the most critical warning in December 2024.
29. Provide the forecasted water availability for 2025 for the Crocodile River.
30. Which country has a well-distributed network of rainfall measuring gauge stations in the Limpopo Basin?